\documentclass[final]{cvpr}

\usepackage{times}
\usepackage{epsfig}
\usepackage{graphicx}
\usepackage{amsmath}
\usepackage{amssymb}
\usepackage{makecell}

\usepackage{color}
\usepackage{bm}
\usepackage{multirow}
\usepackage{graphicx} 
\usepackage{adjustbox}
\usepackage{booktabs,makecell}

\def\eg{\emph{e.g.}}
\def\etal{\emph{et al.}}
\def\ie{\emph{i.e.}}

\usepackage[pagebackref=true,breaklinks=true,colorlinks,bookmarks=false]{hyperref}

\def\cvprPaperID{8225} 
\def\confYear{CVPR 2021}

\begin{document}

\title{
T2VLAD: Global-Local Sequence Alignment for Text-Video Retrieval
}

\author{Xiaohan Wang$^{1,2}$\thanks{Work done during an internship at Baidu Research.}
\quad Linchao Zhu$^{3}$
\quad Yi Yang${^3}$ \\
$^1$Zhejiang University\quad $^2$Baidu Research \quad $^3$ReLER, University of Technology Sydney  \\
{\tt\small wxh1996111@gmail.com}
\quad{\tt\small Linchao.Zhu@uts.edu.au}
\quad {\tt\small Yi.Yang@uts.edu.au}
}

\maketitle

\pagestyle{empty}
\thispagestyle{empty}
\begin{abstract}
Text-video retrieval is a challenging task that aims to search relevant video contents based on natural language descriptions. 
The key to this problem is to measure text-video similarities in a joint embedding space.
However, most existing methods only consider the global cross-modal similarity and overlook the local details. Some works incorporate the local comparisons through cross-modal local matching and reasoning. These complex operations introduce tremendous computation.
In this paper, we design an efficient global-local alignment method. The multi-modal video sequences and text features are adaptively aggregated with a set of shared semantic centers. The local cross-modal similarities are computed between the video feature and text feature within the same center. This design enables the meticulous local comparison and reduces the computational cost of the interaction between each text-video pair.
Moreover, a global alignment method is proposed to provide a global cross-modal measurement that is complementary to the local perspective. 
The global aggregated visual features also provide additional supervision, which is indispensable to the optimization of the learnable semantic centers.
We achieve consistent improvements on three standard text-video retrieval benchmarks and outperform the state-of-the-art by a clear margin. 

\end{abstract}

\section{Introduction}

Video is one of the most informative media due to the abundant multi-modal content and temporal dynamics.
Text-video retrieval systems enable humans to search videos with a simple and natural interaction approach.
Recently, some efforts have been made in building retrieval systems with complex text inputs \cite{chen2020fine,gabeur2020multi}, \eg, retrieving contents of ``a group of men inspect and test a brand new yellow car''.
This is more applicable as the users could search content based on more detailed descriptions.

\begin{figure}[t]
\center
\includegraphics[width=0.8\linewidth]{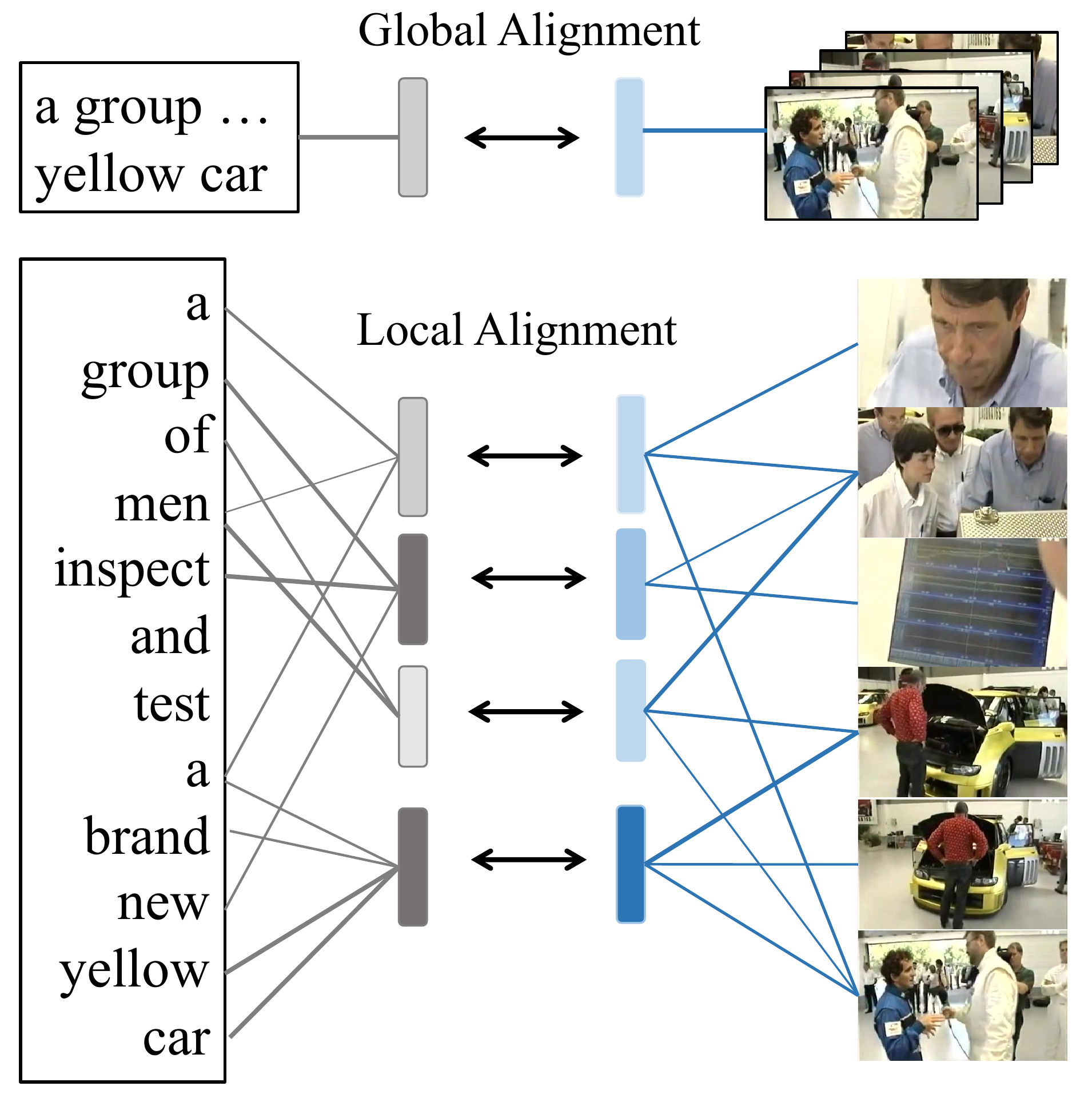}
\caption{Global alignment gives a comprehensive similarity measurement between texts and videos.
Local alignment provides fine-grained comparisons by computing the similarities between the local text-video features from the same semantic centers.
}
\label{fig:overview}  
\end{figure}

One of the promising directions to enable cross-modal video retrieval is to measure text-video similarities using metric learning \cite{xing2003distance,faghri2017vse}. In this case, the common practice is to embed both descriptions and videos into a joint embedding space. 
Most existing works~\cite{miech2018learning}~\cite{dong2019dual}~\cite{Liu2019a}~\cite{gabeur2020multi} encode the descriptions and video content to global representations and compare their similarities from a global perspective.
These methods focus on the learning of effective language and video representations but overlook the fine-grained semantic alignment. For instance, Gabeur \etal~\cite{gabeur2020multi} leveraged a multi-modal transformer to enhance the valuable cross-modal interaction to generate more discriminative video features.
Some other works~\cite{chen2020fine,lee2018stacked,wu2019dual} leveraged complex cross-modal matching operations to exploit the local details and align multiple semantic cues.
Chen \etal~\cite{chen2020fine} proposed a hierarchical graph reasoning model to capture both global events and local actions through local graph matching. 
They manually designed three levels of semantics, including events, actions, and entities.
However, these methods require a high computational cost due to the expensive pairwise matching operation.

In this paper, we propose an efficient global-local sequence alignment method for text-video retrieval.
In the \textbf{local perspective}, we aim to utilize a number of learnable semantic topics to jointly summarize both texts and videos. Instead of parsing text descriptions to a hierarchical semantic role graph \cite{chen2020fine}, it is hoped that these semantic topics could be discovered and automatically learned during the end-to-end training.
We further share the weights of text topics and video topics to offer a joint topic representation learning and to reduce the semantic gap between text and video data.
To achieve local alignment, we minimize the distance between the grouped text feature and the corresponding grouped video features within the same topics. 
In the \textbf{global perspective}, the multi-modal video sequences are aggregated temporally within each modality. The global similarity is computed between the aggregated video features and global text features. The global alignment not only serves as a complementary measurement to local alignment but also provides additional supervision for the learnable semantic topics.
 
We implement the idea of local semantic topic alignment with the help of a NetVLAD operation \cite{arandjelovic2016netvlad}. 
In NetVLAD, the learnable centers are regarded as ``visual words'' of the input data, which can be readily utilized as latent semantic topics on our cross-modal video retrieval task.
For both text and video modalities, we use NetVLAD operations to obtain an aggregated feature for each topic, where the topic centers are \textbf{shared} between the two modalities.
The text features and video features are softly assigned to topics based on their corresponded similarities.
Without complex graph operations \cite{chen2020fine} and multi-layer transformers \cite{gabeur2020multi}, we surprisingly find that our collaborative encoding method, namely Text-to-Video VLAD (T2VLAD), could boost the retrieval performance on various datasets. The contribution of this paper can be summarized as below:
\begin{itemize}
\item First, we propose to automatically learn text-and-video semantic topics and re-emphasize the importance of local semantic alignment between texts and videos for better cross-modal retrieval. 

\item Second, we introduce an effective strategy to locally align text inputs and video inputs.
Based on the success of NetVLAD encoding \cite{arandjelovic2016netvlad}, we propose a T2VLAD encoding for cross-modal retrieval, where we exploit shared centers to reduce the semantic gap between texts and videos instead of the complex pairwise local matching operation.

\item Third, we demonstrate significant improvements of T2VLAD on three standard text-video retrieval benchmarks, \ie, MSRVTT \cite{xu2016msr}, ActivityNet Captions \cite{krishna2017dense}, and {LSMDC} \cite{rohrbach2015dataset}.
Notably, we outperform a HowTo100M-pretrained \cite{miech2019howto100m} multi-modal transformer \cite{gabeur2020multi} with 2.9\% gain (Rank@1) on MSRVTT without any additional data. 
\end{itemize}

\section{Related Work}
\paragraph{Text-Video Retrieval.}
There are increasing interests in advancing text-video retrieval performance \cite{pan2016jointly,fan2020person,chen2020fine,gabeur2020multi}.
Compared to text-image retrieval \cite{faghri2017vse,kiros2014unifying,karpathy2015deep}, text-video retrieval is more challenging that requires the understanding of temporal dynamics and complicated text semantics.
A few works~\cite{pan2016jointly,mithun2018learning} focus on visual semantic embedding learning for text and video joint modeling.
Mithun \etal~\cite{mithun2018learning} leveraged a simple text-image embedding method~\cite{faghri2017vse} to improve the training strategy with hard negative mining, and incorporated multi-modal features (RBG, motion, and audio) to enrich the video representations.
Dong \etal~\cite{dong2019dual} proposed dual-encoding network with multiple levels of features for text-video retrieval, \ie, features obtained by mean pooling, bi-directional Gated Recurrent Unit and Convolution Layers. Yu \etal~\cite{yu2018joint} proposed a joint fusion model using Long Short-Term Memory for temporal sequential information encoding between videos and texts. 
Liu \etal~\cite{Liu2019a} further utilize all modalities that can be extracted from videos such as
speech contents and scene texts for video encoding. 
Miech \etal~\cite{miech2018learning} introduced a strong joint embedding using mixture-of-expert features, which are later utilized in \cite{gabeur2020multi}.

\paragraph{Language Representation Learning.}
Language representations are usually learned using sequence encoders, \eg, Long Short-Term Memory \cite{hochreiter1997long}, Gated Recurrent Unit \cite{chung2014empirical}.
Recently, with the success of BERT model \cite{devlin2018bert} in contextual text representation learning using multi-layer transformer architectures~\cite{vaswani2017attention}, many vision-and-language works \cite{gabeur2020multi,sun2019videobert,zhu2020actbert} leveraged pre-trained BERT features to enhance the language representation capability. Similar to \cite{gabeur2020multi}, we use the BERT model during text-video retrieval and the model is fine-tuned during our end-to-end cross-modal retrieval training.

\paragraph{VLAD Encoding.}
VLAD~\cite{jegou2010aggregating} and NetVLAD~\cite{arandjelovic2016netvlad} have achieved great impacts in aggregating discriminative features for video classification \cite{girdhar2017actionvlad,xu2015discriminative}, video retrieval \cite{miech2018learning}, person re-identification \cite{zheng2017sift}. NetVLAD is an end-to-end differentiable layer that could be readily plugged into many existing models.
These works usually leverage the NetVLAD layer as a discriminative feature learner for downstream tasks. However, in this paper, we leverage NetVLAD in text-video local similarity matching and introduce a local alignment loss to reduce the gap of locally learned features from texts and videos. We do not conduct classification upon the obtained aggregated features, but apply local alignment between the text and video features.

\section{Method}

\begin{figure*}[t]
\center
\includegraphics[width=1.0\linewidth]{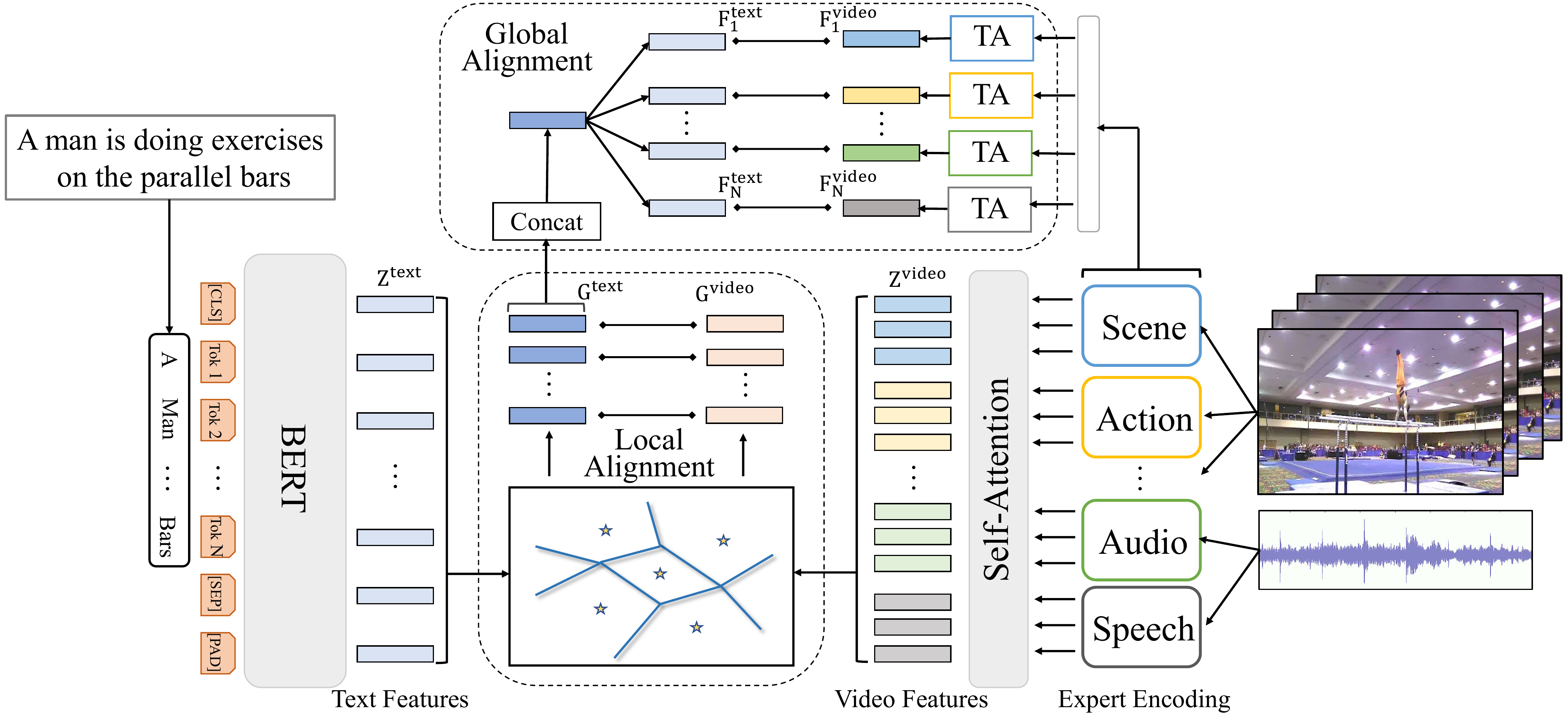}

\caption{Our T2VLAD framework. ``TA'' indicates temporal aggregation. 
Given a text-video pair, we leverage multiple experts to extract the local video features corresponding to each modality. A BERT model is utilized to extract contextual word features. We feed all the video features from different experts to a self-attention layer to enhance the features based on cross-modal relations. The output video features and text features are assigned to a set of shared centers. We aggregate the local features based on the assignments and generate the locally aligned features for both video and text to compute a local video-text similarity. We develop a global alignment scheme in which the video features from each expert are aggregated to a global feature to calculate a similarity with the projected global text feature.
}
\label{fig:framework}  
\end{figure*}

\subsection{Overview}
We propose Text-to-Video VLAD (T2VLAD) for cross-modal retrieval, which aligns text and video features in a global and local perspective.
Given a text-video pair, our goal is to encode it into a joint feature space to measure the similarity.  
As shown in Fig.~\ref{fig:framework}, we leverage multiple experts to extract the local video features corresponding to each modality (Section~\ref{sec:vid}). The BERT model is utilized to extract contextual word features (Section~\ref{sec:text}). After that, we feed all the video features from different experts to a self-attention layer to enhance the features based on cross-modal relations. The output video features and text features are assigned to a set of cluster centers, which are shared between text encoding and video encoding. We aggregate the local features based on the assignments and generate the locally aligned features for both video and text to compute a local video-text similarity (Section~\ref{sec:local}). 
To provide additional supervision on the local alignment and introduce complementary information, we develop a global alignment scheme (Section~\ref{sec:global}).

\subsection{Video Representations}\label{sec:vid}
Compared to image data, videos are more complex and contain richer information such as motion, audio and speech.
To make full use of the multi-modal information in video data for the text-video retrieval task, we leverage multiple experts~\cite{miech2018learning,Liu2019a,gabeur2020multi} to encode raw videos. 
Specifically, given an input video, 
we leverage $N$ experts $\{\bm{E}^1,\bm{E}^2,\ldots,\bm{E}^N\}$ to extract multi-modal features. Here $\bm{E}^n$ represents the $n$-th expert.
Each expert is pretrained on a particular task to acquire specific knowledge on the corresponding modal.
Our goal is to achieve both local and global alignment for text-video retrieval, so we extract features from each temporal segment.
For each expert, we obtain a set of segment-level video representations, \ie,
$\{\bm{E}^n(\bm{x}_1), \bm{E}^n(\bm{x}_2), \ldots, \bm{E}^n(\bm{x}_T)\}$.
Here $T$ is the number of segments, and $\bm{x}_t$ is the $t$-th segment from a video.
We leverage the following two operations to further process the segment-level multi-expert features for the subsequent global-local alignment.

First, we introduce to generate \textbf{global expert features for global alignment}.
We aim to perform temporal aggregation for each expert to generate global expert features. 
There are a few existing temporal aggregation operations to obtain a global vector, \eg, temporal convolution networks~\cite{lea2017temporal}, Transformers~\cite{vaswani2017attention} and NetVLAD~\cite{arandjelovic2016netvlad}. 
For simplicity, we leverage a max-pooling operation without additional parameters. This simple operation works well in our experiments. The temporal-aggregated features are projected to the same dimension for the subsequent clustering.
Following~\cite{miech2018learning}, we then enhanced the features by a self-gating mechanism. 
Consequently, we obtain a set of global expert features $\{\bm{F}^{video}_1,\bm{F}^{video}_2,\ldots,\bm{F}^{video}_N\}$, where $N$ is the number experts.

Second, we use one self-attention layer to \textbf{fuse multi-expert features for local alignment}.
We first employ a fully-connected layer for each expert to project different expert features to a $C$-dimensional embedding space. 
We then concatenate the features from all experts to generate the local features $\bm{Z}^{video} = \{\bm{z}^{video}_1, \bm{z}^{video}_2, \ldots, \bm{z}^{video}_M\}$,
where $M$ is the number of features from all experts. We further explore the relations among the multi-modal features with self-attention mechanism. 
This design is similar to \cite{gabeur2020multi} but has two differences: (1) We only use an one-layer transformer encoder~\cite{vaswani2017attention} instead of the multi-layer transformer with pre-aggregation and position encoding as in~\cite{gabeur2020multi}. Thus, our module introduces fewer parameters and is more computationally efficient; (2) We aim to maintain the locality of the input features while \cite{gabeur2020multi} generates aggregated expert features for the subsequent text-to-video matching. The output feature $\bm{Z}^{video}$ of this process has the same length as the input features.

\subsection{Text Representations}~\label{sec:text}
The BERT model~\cite{devlin2018bert} has shown great generalization capabilities in language feature encoding. 
We leverage a pre-trained BERT model to fairly compared to \cite{gabeur2020multi}.
The BERT model extracts the contextual word embeddings for each text input. 
The input sentences are tokenized and padded to be a fixed-length sequence. The fixed-length sequence is the input to the BERT model.
We add special tokens like ``[CLS]'' and ``[SEP]'' to indicate the start and the end of the sentence. The features can be computed as $\bm{Z}^{text}= \Phi^{BERT}(S)$, where $\Phi^{BERT}$ is the BERT model, $S$ is the input tokens. 
$\bm{Z}^{text} = \{\bm{z}^{text}_1, \bm{z}^{text}_2, \ldots, \bm{z}^{text}_B\}$, where $B$ is the sequence length.
The BERT model $\Phi^{BERT}$ is optimized with the other modules in our framework in an end-to-end manner.
It provides powerful text modeling capacity.
Different from video encoding, the global features for text are extracted jointly with local representations for the subsequent T2VLAD module.

\subsection{Local Alignment}~\label{sec:local}
After the aforementioned text encoding and video encoding, we obtain $B$ local contextual word embeddings $\bm{Z}^{text}$ and $M$ video local features $\bm{Z}^{video}$ for each input text-video pair.
These features contain abundant information about the input sentences and videos. However, the direct comparisons between the two types of features are not feasible because they are not well-aligned. Moreover, the local video features $\bm{Z}^{video}$ are from different modalities. The domain gaps increase the difficulties of the local alignment.
Intuitively, if we select and aggregate the local text features and video features on the same topic and then compare their similarities, the measurement would become more precise. Motivated by this spirit, we propose Text-to-Video VLAD (T2VLAD) to cluster the local features from multiple modalities with shared centers. 
These centers provide shared semantic topics which can bridge the gaps among different modalities. 
Inspired by \cite{arandjelovic2016netvlad}, these centers can be learned jointly with the whole network, and the feature clustering can be performed on-the-fly.

Specifically, we learn $K+1$ $C$-dimensional shared cluster centers $\{\bm{c}_1,\bm{c}_2,\ldots,\bm{c}_K,\bm{c}_{K+1}\}$. Here the $K$ centers are for local alignment and the additional center is for background information removal.
The design of the background center shares the same spirit of~\cite{zhong2018ghostvlad} to discard noise information.
We follow \cite{arandjelovic2016netvlad} to calculate the similarities between each local feature and the cluster centers using dot-product. This step computes assignments on the corresponding clusters.
We start with the encoding of video features. Given a local video feature $\bm{z}^{video}_i$, its assignments to $j$-th cluster can be generated as follows,
\begin{align}
    a_{i,j} = \frac{exp(\bm{z}^{video}_i \bm{c}_j^{\mathsf{T}} +b_j)}{\sum_{k=1}^{K+1} exp(\bm{z}^{video}_i \bm{c}_k^{\mathsf{T}} + b_k)},
\end{align}
where $b_j$ is a learnable bias term. In practice, one can replace the bias term with a batch normalization layer \cite{ioffe2015batch} which normalizes and shifts the activation by two built-in learnable parameters.
Then the aggregated residual feature on each centers can be obtained,
\begin{align}
    \bm{g}^{video}_j =  \text{normalize}( \sum_{i=1}^M a_{i,j}(\bm{z}^{video}_i - \bm{c'}_j)),
\end{align}
where the $\bm{c'}_j$ is trainable weights that have the same size as $\bm{c}_j$, and ``normalize'' indicates a $\ell_2$-normalization operation. The design of introducing two centers for each cluster has been proposed in~\cite{arandjelovic2016netvlad} to increase the adaptation capability of the NetVLAD layer.
We obtain a set of aggregated video feature $\bm{G}^{video}=\{\bm{g}^{video}_1,\bm{g}^{video}_2,\ldots,\bm{g}^{video}_K\}$.
Each feature in $\bm{G}^{video}$ is the aligned local feature for the video.
Note that the aggregated feature on the background center is abandoned and not involved in the following similarity measurement. 

The aggregated text features can be calculated in the same way using the shared cluster centers.
\begin{align}
    \bm{g}^{text}_j =  \text{normalize} (\sum_{i=1}^B \frac{exp(\bm{z}^{text}_i \bm{c}_j^{\mathsf{T}} +b_j)}{\sum_{k=1}^{K+1} exp(\bm{z}^{text}_i \bm{c}_k^{\mathsf{T}} + b_k)}(\bm{z}^{text}_i - \bm{c'}_j)),
\end{align}
where $\bm{z}^{text}_i$ is the local word embedding in $\bm{Z}^{text}$.
We can obtain the final local feature $\bm{G}^{text}=\{\bm{g}^{text}_1,\bm{g}^{text}_2,\ldots,\bm{g}^{text}_K\}$
for the text sequence. 
Since the local feature assignment and aggregation for video and text share the same centers, the final features $\bm{G}^{video}$ and $\bm{G}^{text}$ can be aligned effectively. 
We utilize cosine distance to measure the local similarity between the final video and 
text features $s_{local} = dist(\bm{G}^{video},\bm{G}^{text})$.

\subsection{Global Alignment}~\label{sec:global}
We introduce global alignment for two reasons. First, the global features for text-video pairs are more comprehensive and 
complementary to local features. Second, the elaborate local alignment with trainable centers can be difficult to be optimized when lacking auxiliary supervision, especially when the video features consist of multi-modal information.

Therefore, we alleviate the optimization difficulty in global alignment by aggregating and transforming the video feature from each expert independently. Meanwhile, we utilize the concatenation of local text features $\bm{G}^{text}$ to generate the expert-specific global text representations $\{\bm{F}^{text}_1,\bm{F}^{text}_1,\ldots,\bm{F}^{text}_N\}$. And each feature is then used to compute the similarity with the corresponding video expert feature. Following~\cite{miech2018learning}, we compute the global text-video similarity as a weighted sum of cosine distances between each global video expert feature and corresponding text feature.
Formally, the global similarity is calculated as follows,
\begin{align}
    s_{global} = \sum_{i = 1}^{N}w_i * dist(\bm{F}^{text}_i,\bm{F}^{video}_i),
\end{align}
where $w_i$ is the weight for the $i$-th expert. The weights are generated from the text representation $\bm{G}^{text}$ by a linear projection with a softmax normalization. We utilize the text-video similarity $s = \frac{1}{2}(s_{global} + s_{local})$ to obtain a simple bi-directional max-margin ranking loss on both text-to-video and video-to-text retrieval tasks, following \cite{miech2018learning,gabeur2020multi}. We refer the reader to \cite{miech2018learning,gabeur2020multi} for detailed descriptions.

\begin{table*}[t]
\small
\setlength{\tabcolsep}{7pt}
\centering
\begin{tabular}{l|c|cccc|cccc}
\hline
\multirow{2}{*}{Method} & \multirow{2}{*}{Split} &
\multicolumn{4}{c|}{Text $\rightarrow$ Video} & \multicolumn{4}{c}{Video $\rightarrow$ Text}  \\ 
&  & R@1$\uparrow$ & R@5$\uparrow$ & R@10$\uparrow$ & MdR$\downarrow$ & R@1$\uparrow$ & R@5$\uparrow$ & R@10$\uparrow$ & MdR$\downarrow$ \\ \hline
JSFusion~\cite{yu2018joint} & 1k-A& 10.2& 31.2 & 43.2& 13  & - & - & -& -   \\
HT~\cite{miech2019howto100m} & 1k-A& 14.9 & 40.2 & 52.8 & 9 & - & - &- & -    \\
CE~\cite{Liu2019a} & 1k-A& 20.9 & 48.8 & 62.4 & 6  & 20.6 & 50.3 & 64.0 & 5.3  \\
MMT~\cite{gabeur2020multi} & 1k-A& 24.6 & 54.0 & 67.1 & 4 &  24.4 & 56.0 & 67.8 & 4   \\
{MMT + HT pretrain \cite{gabeur2020multi}} & 1k-A& 26.6 & 57.1 & 69.6 & 4  & 27.0 & 57.5 & 69.7 & 3.7  \\
\hline
Our T2VLAD & 1k-A & \textbf{29.5} & \textbf{59.0} & \textbf{70.1} & \textbf{4}  & \textbf{31.8} & \textbf{60.0} & \textbf{71.1} & \textbf{3} \\
\hline
\hline

MEE~\cite{miech2018learning} & 1k-B & 13.6 & 37.9 & 51.0 & 10  & - & - & - & -  \\
JPose~\cite{wray2019fine} & 1k-B & 14.3 & 38.1 & 53.0 & 9  & 16.4 & 41.3 & 54.4 & 8.7  \\
MEE-COCO~\cite{miech2018learning} & 1k-B & 14.2 & 39.2 & 53.8 & 9  & - & - & - & - \\
CE~\cite{Liu2019a} & 1k-B& 18.2 & 46.0 & 60.7 & 7  & 18.0 & 46.0 & 60.3 & 6.5  \\
MMT~\cite{gabeur2020multi}& 1k-B & 
20.3 & 49.1 & 63.9 & 6  & 21.1 & 49.4 & 63.2 & 6 \\
\hline
Our T2VLAD & 1k-B & \textbf{26.1} & \textbf{54.7} & \textbf{68.1} & \textbf{4}  & \textbf{26.7} & \textbf{56.1} & \textbf{70.4} & \textbf{4}  \\ 
\hline
\hline
VSE~\cite{mithun2018learning} & Full & 5.0 & 16.4 & 24.6 & 47 & 7.7 & 20.3 & 31.2 & 28 \\
VSE++~\cite{mithun2018learning} & Full & 5.7 & 17.1 & 24.8 & 65  & 10.2 & 25.4 & 35.1 & 25  \\
Mithun~\etal~\cite{mithun2018learning}  & Full & 7.0 & 20.9 & 29.7 & 38 & 12.5 & 32.1 & 42.4 & 16 \\
W2VV~\cite{dong2018predicting} & Full & 6.1 & 18.7 & 27.5 & 45  & 11.8 & 28.9 & 39.1 & 21  \\
Dual Enc.~\cite{dong2019dual} & Full & 7.7 & 22.0 & 31.8 & 32 & 13.0 & 30.8 & 43.3 & 15  \\
HGR~\cite{chen2020fine}& Full &  9.2 & 26.2 & 36.5 & 24 & 15.0 & 36.7 & 48.8 & 11  \\
E2E~\cite{miech2020end} & Full & 9.9 & 24.0 & 32.4 & 29.5  & - & - & - & -  \\
CE~\cite{Liu2019a} & Full & 10.0 & 29.0 & 41.2 & 16 & 15.6 & 40.9 & 55.2& 8.3 \\
\hline
Our T2VLAD & Full & \textbf{12.7} & \textbf{34.8} & \textbf{47.1} & \textbf{12} & \textbf{20.7} & \textbf{48.9} & \textbf{62.1} & \textbf{6}  \\ \hline

\end{tabular}
\caption{The comparison with the state-of-the-art methods on the MSRVTT~\cite{xu2016msr} dataset.}
\label{tab:msrvtt}
\end{table*}

\begin{table*}[t]
\small
\setlength{\tabcolsep}{11pt}
\centering
\begin{tabular}{l|cccc|cccc}
\hline
\multirow{2}{*}{Method} &
\multicolumn{4}{c|}{Text $\rightarrow$ Video} & \multicolumn{4}{c}{Video $\rightarrow$ Text}  \\ 
 & R@1 $\uparrow$ & R@5 $\uparrow$ & R@50 $\uparrow$ & MdR $\downarrow$  & R@1 $\uparrow$ & R@5 $\uparrow$ & R@50 $\uparrow$ & MdR $\downarrow$ \\ \hline
 
FSE~\cite{zhang2018cross} & 18.2 & 44.8 & 89.1 & 7   & 16.7 & 43.1 & 88.4 & 7  \\
CE~\cite{Liu2019a} & 18.2 & 47.7 & 91.4 & 6 & 17.7 & 46.6 & 90.9 & 6 \\
HSE~\cite{zhang2018cross} & 20.5 & 49.3 & - & -  & 18.7 & 48.1 & - & - \\
MMT~\cite{gabeur2020multi} & 22.7 & 54.2 & 93.2 & 5 & 22.9 & 54.8 & 93.1 & 4.3  \\
\hline
Ours & 
\textbf{23.7} & \textbf{55.5} & \textbf{93.5} & \textbf{4}   & \textbf{24.1} & \textbf{56.6} & \textbf{94.1} & \textbf{4} \\ \hline

\end{tabular}
\caption{The comparisons with the state-of-the-art methods on the ActivityNet Captions dataset.}
\vspace{-2mm}
\label{tab:acnet}
\end{table*}

\begin{table*}[t]
\small
\setlength{\tabcolsep}{10pt}
\centering
\begin{tabular}{l|cccc|cccc}
\hline
\multirow{2}{*}{Method} &
\multicolumn{4}{c|}{Text $\rightarrow$ Video} & \multicolumn{4}{c}{Video $\rightarrow$ Text}  \\ 
 & R@1 $\uparrow$ & R@5 $\uparrow$ & R@50 $\uparrow$ & MdR $\downarrow$  & R@1 $\uparrow$ & R@5 $\uparrow$ & R@50 $\uparrow$ & MdR $\downarrow$  \\ \hline
 
CT-SAN \cite{yu2017end} & 5.1 & 16.3 & 25.2 & 46   & - & - &- & -  \\
JSFusion \cite{yu2018joint} & 9.1 & 21.2 & 34.1 & 36   & - & - &- & -  \\
CCA \cite{klein2015associating} & 7.5 &  21.7 & 31.0 & 33  & - & - &- & -  \\
MEE \cite{miech2018learning} & 9.3 & 25.1 & 33.4 & 27 & - & - &- & - 
\\
MEE-COCO \cite{miech2018learning} & 10.1 & 25.6  & 34.6 &27  & - & - &- & -  \\
CE \cite{Liu2019a} & 11.2 & 26.9& 34.8 & 25.3   & - & - &- & -   \\
MMT~\cite{gabeur2020multi} & 13.2&  29.2& 38.8& 21 & 12.1& 29.3& 37.9& 22.5 \\
\hline
Ours & \textbf{14.3} & \textbf{32.4} & \textbf{42.2} & \textbf{16} & \textbf{14.2} & \textbf{33.5} & \textbf{41.7} & \textbf{17} \\ \hline
\end{tabular}
\caption{The comparison with the state-of-the-art methods on the LSMDC dataset.}
\label{tab:lsmdc}
\end{table*}

\begin{table*}[t]
\small
\setlength{\tabcolsep}{9pt}
\centering
\begin{tabular}{l|cccc|cccc}
\hline
\multirow{2}{*}{Method}  &
\multicolumn{4}{c|}{Text $\rightarrow$ Video} & \multicolumn{4}{c}{Video $\rightarrow$ Text}  \\ 
& R@1$\uparrow$ & R@5$\uparrow$ & R@10$\uparrow$ & MdR$\downarrow$   & R@1$\uparrow$ & R@5$\uparrow$ & R@10$\uparrow$ & MdR$\downarrow$  \\ \hline
Ours w/o Global Alignment & 24.3 & 51.5&  63.4 & 5  & 26.6 & 52.9 & 62.6 & 5 \\

Ours w/o Local Alignment & 22.2 & 49.9 & 64.6 & 6  & 24.0 & 51.7 & 65.6 & 5 \\
Full model & \textbf{29.5} & \textbf{59.0} & \textbf{70.1} & \textbf{4} & \textbf{31.8} & \textbf{60.0} & \textbf{71.1} & \textbf{3}  \\  
\hline
\end{tabular}
\caption{The ablation studies on the MSRVTT~\cite{xu2016msr} dataset to investigate the effectiveness of global-local alignment.}
\label{tab:ablation1}
\end{table*}

\begin{table*}[t]
\small
\setlength{\tabcolsep}{8pt}
\centering
\begin{tabular}{l|cccc|cccc}
\hline
\multirow{2}{*}{Method}  &
\multicolumn{4}{c|}{Text $\rightarrow$ Video} & \multicolumn{4}{c}{Video $\rightarrow$ Text}  \\ 
& R@1$\uparrow$ & R@5$\uparrow$ & R@10$\uparrow$ & MdR$\downarrow$   & R@1$\uparrow$ & R@5$\uparrow$ & R@10$\uparrow$ & MdR$\downarrow$  \\ \hline
Ours w/ only text VLAD & 27.4 & 57.3 & 68.2 & 4 & 27.5 & 57.4 & 69.7 & 4  \\
Ours w/ two separate VLAD & 28.6& 58.1 & 70.4 & 4 & 30.4 & 60.7 & 72.1 & 3 \\
Ours w/ two shared VLAD   & \textbf{29.5} & \textbf{59.0} & 70.1 & \textbf{4} & \textbf{31.8} & 60.0 & 71.1 & \textbf{3}  \\
\hline
\end{tabular}
\caption{The ablation studies on the MSRVTT~\cite{xu2016msr} dataset to investigate the effectiveness of the VLAD encoding.}
\label{tab:ablation2}
\end{table*}

\section{Experiments}
\subsection{Experimental Details}

\noindent\textbf{Dataset.}
We experiment with MSRVTT \cite{xu2016msr}, video-text datasets. 
The \textbf{MSRVTT} dataset contains 10,000 videos. These videos are collected from YouTube using 257 queries from a commercial video search engine. 
We evaluate the performance on three splits.
For the ``1k-A'' split, the train and test are split as introduced in \cite{yu2018joint}.
The ``1k-B'' split is obtained following \cite{miech2018learning}.
Both splits use 9,000 videos for training, and the remaining 1,000 videos are used for testing.
The \textbf{ActivityNet  Captions} dataset \cite{krishna2017dense}  consists  of  20,000  videos. Each video is densely annotated with multiple sentence descriptions. 
The \textbf{LSMDC} dataset \cite{rohrbach2015dataset} consists of 118,081 short video clips. The videos are extracted from 202 long movies. 

\noindent\textbf{Evaluation Metrics.} 
We report the results with the standard video retrieval metrics, \ie,  Rank $K$ (R@K, higher is better),  Median  Rank (MdR, lower is better). 
We report R@1, R@5, and R@10 following \cite{chen2020fine,Liu2019a}.

\noindent\textbf{Multi-Expert Features.}
We use the features provided by \cite{gabeur2020multi} in our experiments.
These features are: {Motion} features from S3D~\cite{xie2018rethinking} trained on the Kinetics dataset. 
 {Audio} features from VGGish model~\cite{hershey2017cnn} trained on YT8M. {Scene} embeddings from DenseNet-161~\cite{huang2017densely} trained on the Places365 dataset~\cite{zhou2017places}. We refer the readers to \cite{gabeur2020multi} for more descriptions of OCR, Face, Speech, and Appearance features. For MSRVTT, we also leverage optical flow features released by \cite{gabeur2020multi}. We do not use Speech features on LSMDC due to feature missing from the released features \cite{gabeur2020multi}.

\noindent\textbf{Implantation Details.}
We train the projection layer and the T2VLAD module from scratch, and no additional data is used. The margin in the ranking loss is set to 0.02 for all datasets. Following~\cite{Liu2019a}, we leverage Ranger optimizer with a weight decay 0.0001.
We initialize the learning rate at 0.0001, and decay by a multiplicative factor 0.9 every 5 epochs. The batch size of the video-text pairs is set to 64.
For text encoding, we use the pretrained BERT model ``BERT-base-uncased'' and fine-tune it with our framework in an end-to-end manner. For video expert encoding, we leverage the pre-extracted expert features provided by~\cite{gabeur2020multi}. We use all 8 experts for the MSRVTT dataset and 6 experts (rgb, audio, ocr, scene, flow and action) for the LSMDC dataset. For ActivityNet Captions, we only use motion and audio experts. The self-attention module used for local video features is implemented by one layer multi-head attention with 4 heads, a dropout probability of 0.1, and a hidden size of 768. The dimension for the common space of both global alignment and local alignment is also set to 768. We set the center size of our T2VLAD to 9 for the short video retrieval dataset (MSRVTT and LSMDC) and 16 for the long video retrieval dataset (ActivityNet Captions).

\subsection{Comparison to State-of-the-art}
\noindent\textbf{MSRVTT.} The results on MSRVTT are shown in Table \ref{tab:msrvtt}. We consistently improve the state-of-the-art on text-to-video retrieval and video-to-text retrieval across all three splits. 
MMT \cite{gabeur2020multi} is recently proposed to perform text-video retrieval using multi-modal transformers. It achieved the best performance in the compared methods.
Notably, for text-to-video retrieval, we outperform MMT \cite{gabeur2020multi} with 5.8\% gain on the R@1 metric on the 1k-B split (20.3\% \vs 26.1\%).
A 5.6\% improvement on R@1 (1k-B split) is also obtained compared to MMT \cite{gabeur2020multi} for video-to-text retrieval (21.1\% \vs 26.7\%). These results demonstrate the benefits of our T2VLAD in cross-modal retrieval tasks. Notably, we obtain consistent improvements over ``MMT + HT pretrain'' \cite{gabeur2020multi} on the 1k-A split. ``MMT + HT pretrain'' is pre-trained on a large-scale instructional video dataset, \ie, HowTo100M, containing more than one hundred million video clips with machine-generated descriptions. Pre-training on HowTo100M significantly improves the performance of MMT across all evaluation metrics. 
T2VLAD does not leverage additional training videos, but we outperform ``MMT + HT pretrain'' on split 1k-A with a clear margin across all metrics. For instance, on text-to-video retrieval, T2VLAD outperforms ``MMT + HT pretrain'' by 2.9\% at R@1. These results demonstrate that the benefit of the global-local alignment using T2VLAD.

The efficiency of our method is demonstrated by calculating inference time for 1k videos and 1k text queries from MSRVTT on a single V100 GPU. Our video encoding module (except expert encoding) only takes 0.4s for process 1k videos while MMT takes 1.1s. This shows the superiority of our efficient T2VLAD design.

\noindent\textbf{ActivityNet Captions.}
ActivityNet Captions consists of long videos and the captions contain several sentences. The results on this dataset are shown in Table \ref{tab:acnet}. The compared baselines include HSE \cite{chen2020fine}, CE~\cite{Liu2019a}, HSE~\cite{zhang2018cross}, and MMT~\cite{gabeur2020multi}. HSE~\cite{zhang2018cross} leverages a hierarchical sequence embedding and MMT incorporates multi-layer transformers for strong video feature learning. We consistently improve MMT over all benchmark metrics, which demonstrates the effectiveness of T2VLAD on long-term text-video modeling.

\noindent\textbf{LSMDC.} The LSMDC data is collected from movies. The results are shown in Table \ref{tab:lsmdc}.
We observe consistent improvements over MMT. For instance, we achieve 2.1\% improvements on R@1 for video-to-text retrieval. The results show that our T2VLAD is capable of dealing with different videos from different domains.

\begin{figure*}[!ht]
\center
\includegraphics[width=1.0\linewidth]{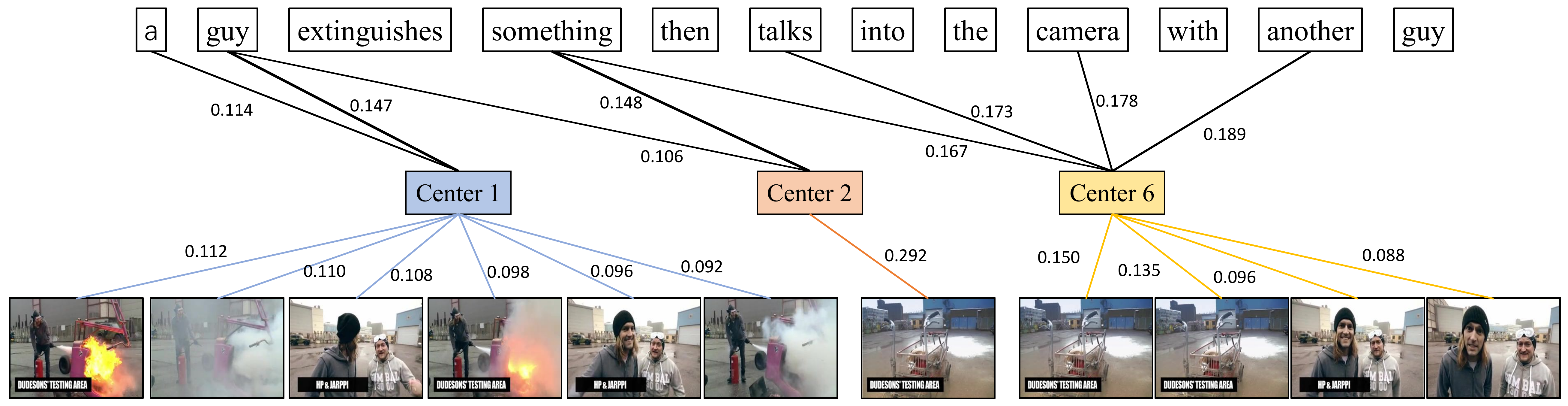}
\caption{Visualization of the assignment weights. We take Video 7060 in the MSRVTT 1K-A test set as an example. We plot the top text assignments to the three centers as black lines and put the assignment values next to the line. 
The Top-10 frames (the padding features have been removed.) correspond to the appearance features assigned to the centers are shown at the bottom. 
}
\vspace{-2mm}
\label{fig:weight}  
\end{figure*}

\begin{figure*}[!ht]
\center
\includegraphics[width=0.9\linewidth]{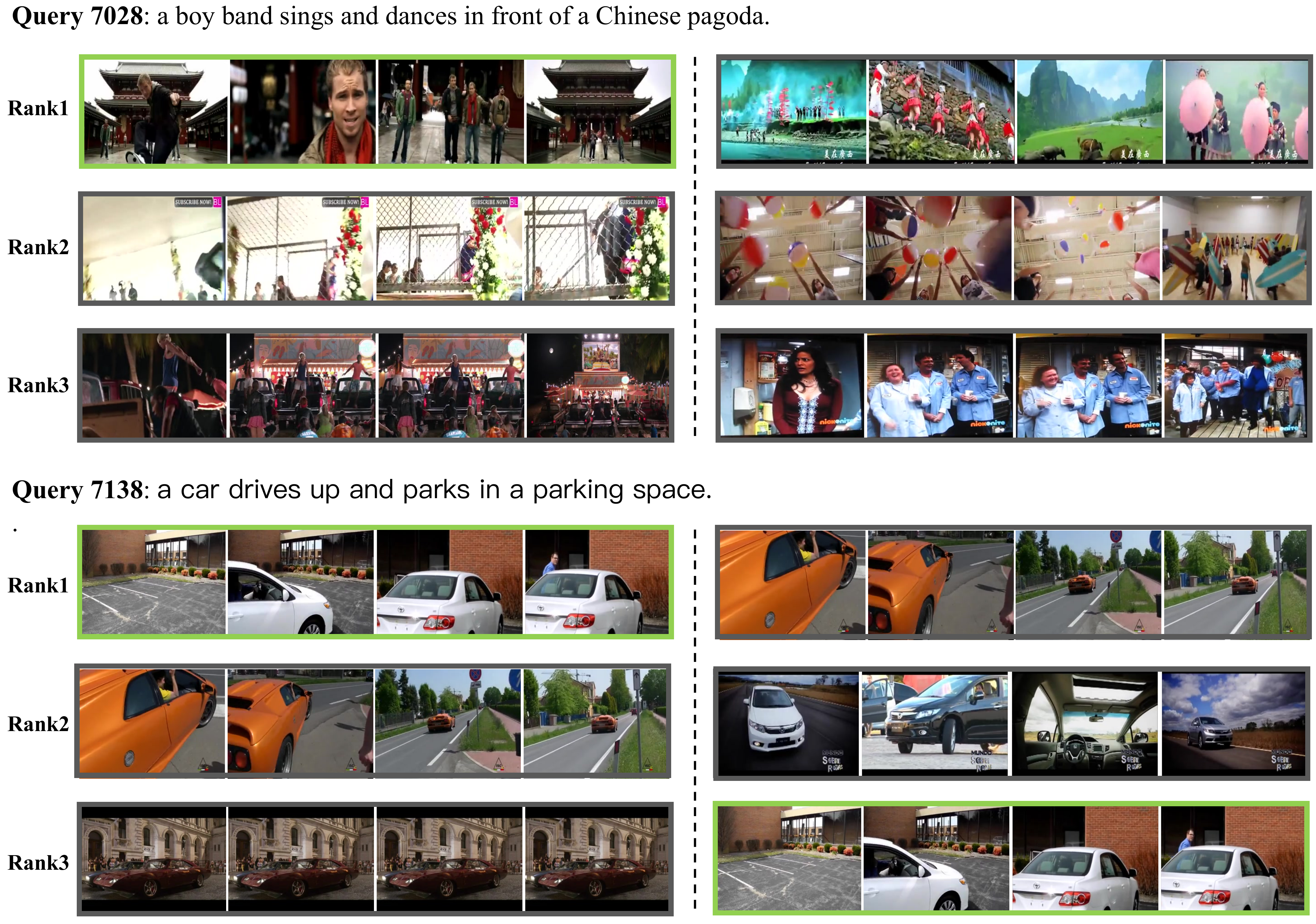}
\caption{The text-video retrieval results on the MSRVTT 1K-A test set. The left are the videos ranked by our T2VLAD, and the right are the results from the model with only global alignment.}
\vspace{-2mm}
\label{fig:result_vis}  
\end{figure*}

\subsection{Ablation Study}

\noindent\textbf{The effectiveness of the global-local alignment.} In Table~\ref{tab:ablation1}, we show the results of only using the single alignment of our model. To implement the model without local alignment, we follow ~\cite{gabeur2020multi} to utilize the ``[CLS]'' output of the BERT model as the global text representation. 
When we remove the local alignment branch and only train the global alignment, the test performance drops a lot compared to
the results of our full model. This proves our local alignment is crucial for the cross-modal retrieval task.
When we remove the global alignment and only train the local alignment, the loss can not converge. It demonstrates the importance of global alignment for providing additional supervision for the optimization of the local alignment. 
We show the results of removing the global alignment only at test time, \ie, ``Ours w/o Global Alignment'' in Table~\ref{tab:ablation1}. Compared to the full model, the results drop by 5.2\% on R@1 for text-to-video retrieval. It demonstrates that the global feature is complementary to the local information.

\noindent\textbf{The effectiveness of collaborative VLAD.} 
In ``Ours w/ only text VLAD'', we replace the shared NetVLAD layer for local video feature encoding with a max-pooling operation and then project the feature to the same dimension of text local features. This model achieves lower performance than our T2VLAD, showing the importance of joint VLAD encoding.
In ``Ours w/ two separate VLAD'', we do not perform center sharing between text feature encoding and video feature encoding. The VLAD centers are learned separately. The results show that our strategy of sharing centers outperform ``Ours w/ two separate VLAD'' especially for text-to-video retrieval.
This demonstrates that our center sharing idea is beneficial to reduce the semantic gap between text and video data.

\subsection{Qualitative Results}
\noindent\textbf{Visualization of the assignments.}
The text local features and the local video features are assigned to a set of shared centers in our T2VLAD. We expect the aggregated text feature and video feature on the same center to share a similar topic. In Fig.~\ref{fig:weight},
we illustrate the text assignments and video appearance feature assignments on three centers. 
The video is ranked first in the text retrieved results.
We show the video frames corresponding to the appearance features that are assigned to the certain center.
As shown in Fig.~\ref{fig:weight}, the text feature with the highest assignment on Center 1 is the feature of ``guy''. All the frames that have been assigned to Center 1 also contain the appearance information of ``guy''. The text with the highest assignment on Center 2 is ``something'',  and the only frame assigned to the center is about the ``something'' in the video. On Center 6, the text ``something'', ``talks'', ``camera'' and ``another'' all have high assignments. And the frames assigned to the center contain these content. Interestingly, the most salient word ``extinguishes'' in the human view, always has a low assignment value on all centers. This is because the limited training data is not enough to enable the understanding of a low-frequency word.
The assignment visualization verifies that our T2VLAD can achieve adequate local alignment for text-to-video retrieval.

\noindent\textbf{Visualization of the text-to-video results.}
We show two examples of the videos retrieved by our method and the model without the local alignment branch. As shown in Fig.~\ref{fig:result_vis}, the two query sentences consist of multiple semantic topics. Our T2VLAD successfully retrieves the ground-truth video while the model without local alignment returns several videos that are somewhat relevant to the query sentence but are not precise. In the second example, our T2VLAD achieves a better alignment between the text and videos on the local semantic cue ``parks''. These results demonstrate that our T2VLAD can align multiple semantic cues effectively.

\section{Conclusion}
In this paper, we introduce an end-to-end text-video sequence alignment method. We show that local semantic alignment between texts and videos is critical for high-performance retrieval systems. We achieve the goal of local alignment based on NetVLAD and introduce T2VLAD for collaborative text-video encoding. The results on three standard text-video retrieval benchmarks clearly demonstrate the effectiveness of our method. The visualization results also validate our motivation for joint semantic topic learning. In the future, more efforts could be paid to obtain better global video features with end-to-end optimization.

\textbf{Acknowledgement} Thanks to Samuel Albanie for his kind help with dataset processing and to Shizhe Chen for valuable discussions.

{\small
\bibliographystyle{ieee_fullname}
\bibliography{egbib}

\begin{thebibliography}{10}\itemsep=-1pt

\bibitem{arandjelovic2016netvlad}
Relja Arandjelovic, Petr Gronat, Akihiko Torii, Tomas Pajdla, and Josef Sivic.
\newblock Netvlad: Cnn architecture for weakly supervised place recognition.
\newblock In {\em CVPR}, 2016.

\bibitem{chen2020fine}
Shizhe Chen, Yida Zhao, Qin Jin, and Qi Wu.
\newblock Fine-grained video-text retrieval with hierarchical graph reasoning.
\newblock In {\em CVPR}, 2020.

\bibitem{chung2014empirical}
Junyoung Chung, Caglar Gulcehre, KyungHyun Cho, and Yoshua Bengio.
\newblock Empirical evaluation of gated recurrent neural networks on sequence
  modeling.
\newblock {\em arXiv preprint arXiv:1412.3555}, 2014.

\bibitem{devlin2018bert}
Jacob Devlin, Ming-Wei Chang, Kenton Lee, and Kristina Toutanova.
\newblock Bert: Pre-training of deep bidirectional transformers for language
  understanding.
\newblock {\em arXiv preprint arXiv:1810.04805}, 2018.

\bibitem{dong2018predicting}
Jianfeng Dong, Xirong Li, and Cees~GM Snoek.
\newblock Predicting visual features from text for image and video caption
  retrieval.
\newblock {\em IEEE Transactions on Multimedia}, 2018.

\bibitem{dong2019dual}
Jianfeng Dong, Xirong Li, Chaoxi Xu, Shouling Ji, Yuan He, Gang Yang, and Xun
  Wang.
\newblock Dual encoding for zero-example video retrieval.
\newblock In {\em CVPR}, 2019.

\bibitem{faghri2017vse}
Fartash Faghri, David~J Fleet, Jamie~Ryan Kiros, and Sanja Fidler.
\newblock Vse++: Improving visual-semantic embeddings with hard negatives.
\newblock {\em arXiv preprint arXiv:1707.05612}, 2017.

\bibitem{fan2020person}
Hehe Fan and Yi Yang.
\newblock Person tube retrieval via language description.
\newblock In {\em AAAI}, 2020.

\bibitem{gabeur2020multi}
Valentin Gabeur, Chen Sun, Karteek Alahari, and Cordelia Schmid.
\newblock Multi-modal transformer for video retrieval.
\newblock In {\em ECCV}, 2020.

\bibitem{girdhar2017actionvlad}
Rohit Girdhar, Deva Ramanan, Abhinav Gupta, Josef Sivic, and Bryan Russell.
\newblock Actionvlad: Learning spatio-temporal aggregation for action
  classification.
\newblock In {\em CVPR}, 2017.

\bibitem{hershey2017cnn}
Shawn Hershey, Sourish Chaudhuri, Daniel~PW Ellis, Jort~F Gemmeke, Aren Jansen,
  R~Channing Moore, Manoj Plakal, Devin Platt, Rif~A Saurous, Bryan Seybold,
  et~al.
\newblock Cnn architectures for large-scale audio classification.
\newblock In {\em ICASSP}, 2017.

\bibitem{hochreiter1997long}
Sepp Hochreiter and J{\"u}rgen Schmidhuber.
\newblock Long short-term memory.
\newblock {\em Neural Computation}, 1997.

\bibitem{huang2017densely}
Gao Huang, Zhuang Liu, Laurens Van Der~Maaten, and Kilian~Q Weinberger.
\newblock Densely connected convolutional networks.
\newblock In {\em CVPR}, 2017.

\bibitem{ioffe2015batch}
Sergey Ioffe and Christian Szegedy.
\newblock Batch normalization: Accelerating deep network training by reducing
  internal covariate shift.
\newblock {\em arXiv preprint arXiv:1502.03167}, 2015.

\bibitem{jegou2010aggregating}
Herv{\'e} J{\'e}gou, Matthijs Douze, Cordelia Schmid, and Patrick P{\'e}rez.
\newblock Aggregating local descriptors into a compact image representation.
\newblock In {\em CVPR}, 2010.

\bibitem{karpathy2015deep}
Andrej Karpathy and Li Fei-Fei.
\newblock Deep visual-semantic alignments for generating image descriptions.
\newblock In {\em CVPR}, 2015.

\bibitem{kiros2014unifying}
Ryan Kiros, Ruslan Salakhutdinov, and Richard~S Zemel.
\newblock Unifying visual-semantic embeddings with multimodal neural language
  models.
\newblock {\em arXiv preprint arXiv:1411.2539}, 2014.

\bibitem{klein2015associating}
Benjamin Klein, Guy Lev, Gil Sadeh, and Lior Wolf.
\newblock Associating neural word embeddings with deep image representations
  using fisher vectors.
\newblock In {\em CVPR}, 2015.

\bibitem{krishna2017dense}
Ranjay Krishna, Kenji Hata, Frederic Ren, Li Fei-Fei, and Juan Carlos~Niebles.
\newblock Dense-captioning events in videos.
\newblock In {\em ICCV}, 2017.

\bibitem{lea2017temporal}
Colin Lea, Michael~D Flynn, Rene Vidal, Austin Reiter, and Gregory~D Hager.
\newblock Temporal convolutional networks for action segmentation and
  detection.
\newblock In {\em CVPR}, 2017.

\bibitem{lee2018stacked}
Kuang-Huei Lee, Xi Chen, Gang Hua, Houdong Hu, and Xiaodong He.
\newblock Stacked cross attention for image-text matching.
\newblock In {\em ECCV}, 2018.

\bibitem{Liu2019a}
Y. Liu, S. Albanie, A. Nagrani, and A. Zisserman.
\newblock Use what you have: Video retrieval using representations from
  collaborative experts.
\newblock In {\em BMVC}, 2019.

\bibitem{miech2020end}
Antoine Miech, Jean-Baptiste Alayrac, Lucas Smaira, Ivan Laptev, Josef Sivic,
  and Andrew Zisserman.
\newblock End-to-end learning of visual representations from uncurated
  instructional videos.
\newblock In {\em CVPR}, 2020.

\bibitem{miech2018learning}
Antoine Miech, Ivan Laptev, and Josef Sivic.
\newblock Learning a text-video embedding from incomplete and heterogeneous
  data.
\newblock {\em arXiv preprint arXiv:1804.02516}, 2018.

\bibitem{miech2019howto100m}
Antoine Miech, Dimitri Zhukov, Jean-Baptiste Alayrac, Makarand Tapaswi, Ivan
  Laptev, and Josef Sivic.
\newblock Howto100m: Learning a text-video embedding by watching hundred
  million narrated video clips.
\newblock In {\em ICCV}, 2019.

\bibitem{mithun2018learning}
Niluthpol~Chowdhury Mithun, Juncheng Li, Florian Metze, and Amit~K
  Roy-Chowdhury.
\newblock Learning joint embedding with multimodal cues for cross-modal
  video-text retrieval.
\newblock In {\em ICMR}, 2018.

\bibitem{pan2016jointly}
Yingwei Pan, Tao Mei, Ting Yao, Houqiang Li, and Yong Rui.
\newblock Jointly modeling embedding and translation to bridge video and
  language.
\newblock In {\em CVPR}, 2016.

\bibitem{rohrbach2015dataset}
Anna Rohrbach, Marcus Rohrbach, Niket Tandon, and Bernt Schiele.
\newblock A dataset for movie description.
\newblock In {\em CVPR}, 2015.

\bibitem{sun2019videobert}
Chen Sun, Austin Myers, Carl Vondrick, Kevin Murphy, and Cordelia Schmid.
\newblock Videobert: A joint model for video and language representation
  learning.
\newblock In {\em ICCV}, 2019.

\bibitem{vaswani2017attention}
Ashish Vaswani, Noam Shazeer, Niki Parmar, Jakob Uszkoreit, Llion Jones,
  Aidan~N Gomez, {\L}ukasz Kaiser, and Illia Polosukhin.
\newblock Attention is all you need.
\newblock In {\em NeurIPS}, 2017.

\bibitem{wray2019fine}
Michael Wray, Diane Larlus, Gabriela Csurka, and Dima Damen.
\newblock Fine-grained action retrieval through multiple parts-of-speech
  embeddings.
\newblock In {\em ICCV}, 2019.

\bibitem{wu2019dual}
Yu Wu, Linchao Zhu, Yan Yan, and Yi Yang.
\newblock Dual attention matching for audio-visual event localization.
\newblock In {\em ICCV}, 2019.

\bibitem{xie2018rethinking}
Saining Xie, Chen Sun, Jonathan Huang, Zhuowen Tu, and Kevin Murphy.
\newblock Rethinking spatiotemporal feature learning: Speed-accuracy trade-offs
  in video classification.
\newblock In {\em ECCV}, 2018.

\bibitem{xing2003distance}
Eric~P Xing, Michael~I Jordan, Stuart~J Russell, and Andrew~Y Ng.
\newblock Distance metric learning with application to clustering with
  side-information.
\newblock In {\em NeurIPS}, 2003.

\bibitem{xu2016msr}
Jun Xu, Tao Mei, Ting Yao, and Yong Rui.
\newblock Msr-vtt: A large video description dataset for bridging video and
  language.
\newblock In {\em CVPR}, 2016.

\bibitem{xu2015discriminative}
Zhongwen Xu, Yi Yang, and Alex~G Hauptmann.
\newblock A discriminative cnn video representation for event detection.
\newblock In {\em CVPR}, 2015.

\bibitem{yu2018joint}
Youngjae Yu, Jongseok Kim, and Gunhee Kim.
\newblock A joint sequence fusion model for video question answering and
  retrieval.
\newblock In {\em ECCV}, 2018.

\bibitem{yu2017end}
Youngjae Yu, Hyungjin Ko, Jongwook Choi, and Gunhee Kim.
\newblock End-to-end concept word detection for video captioning, retrieval,
  and question answering.
\newblock In {\em CVPR}, 2017.

\bibitem{zhang2018cross}
Bowen Zhang, Hexiang Hu, and Fei Sha.
\newblock Cross-modal and hierarchical modeling of video and text.
\newblock In {\em ECCV}, 2018.

\bibitem{zheng2017sift}
Liang Zheng, Yi Yang, and Qi Tian.
\newblock Sift meets cnn: A decade survey of instance retrieval.
\newblock {\em TPAMI}, 2017.

\bibitem{zhong2018ghostvlad}
Yujie Zhong, Relja Arandjelovi{\'c}, and Andrew Zisserman.
\newblock Ghostvlad for set-based face recognition.
\newblock In {\em ACCV}, 2018.

\bibitem{zhou2017places}
Bolei Zhou, Agata Lapedriza, Aditya Khosla, Aude Oliva, and Antonio Torralba.
\newblock Places: A 10 million image database for scene recognition.
\newblock {\em TPAMI}, 2017.

\bibitem{zhu2020actbert}
Linchao Zhu and Yi Yang.
\newblock Actbert: Learning global-local video-text representations.
\newblock In {\em CVPR}, 2020.

\end{thebibliography}
}

\end{document}